# ACCL: Adversarial constrained-CNN loss for weakly supervised medical image segmentation

Pengyi Zhang
Beijing Institute of Technology
Beijing, China
zhangpybit@gmail.com

Yunxin Zhong
Beijing Institute of Technology
Beijing, China
bityunxinz@gmail.com

Xiaoqiong Li
Beijing Institute of Technology
Beijing, China
aeople@126.com

**Abstract**

Weakly supervised semantic segmentation is attracting significant attention in medical image analysis as only low-cost weak annotations, e.g., point, scribble or box annotations, are required to train CNNs. Constrained-CNN loss is one poplar approach for weakly supervised segmentation by imposing inequality constraints of prior knowledge, e.g., the size and shape of the object of interest, on network's outputs. However, describing the prior knowledge, e.g., irregular shape and unsmooth boundary, in programming language may not be so easy. In this paper, we propose adversarial constrained-CNN loss (ACCL), a new paradigm of constrained-CNN loss methods, for weakly supervised medical image segmentation. In the new paradigm, prior knowledge, e.g., the size and shape of the object of interest, is encoded and depicted by reference masks, and is further employed to impose constraints on segmentation outputs through adversarial learning with reference masks. Unlike pseudo label methods for weakly supervised segmentation, such reference masks are used to train a discriminator rather than a segmentation network, and thus are not required to be paired with specific images. Our new paradigm not only greatly facilitates imposing prior knowledge on network's outputs, but also provides stronger and higher-order constraints, i.e., distribution approximation, through adversarial learning. Extensive experiments involving different medical modalities, different anatomical structures, different topologies of the object of interest, different levels of prior knowledge and weakly supervised annotations with different annotation ratios is conducted to evaluate our ACCL method. Consistently superior segmentation results over the size constrained-CNN loss method have been achieved, some of which are close to the results of full supervision, thus fully verifying the effectiveness and generalization of our method. Specifically, we report an average Dice score of 75.4% with an average annotation ratio of 0.65%, surpassing the prior art, i.e., the size constrained-CNN loss method, by a large margin of 11.4%. Our codes are made publicly available at https://github.com/PengyiZhang/ACCL.

**1. Introduction**

In the recent years, deep convolutional neural network (CNN) has been witnessed as a promising technique for many medical image analysis tasks, e.g., diabetic retinopathy detection [1], skin cancer diagnosis [2], lung disease detection [3] and heart disease risk prediction [4], due to end-to-end learning framework and availability of large-scale labelled samples. In particular, many modern CNN architectures, for instance, UNet [5], VNet[6], ENet [7], and UNet++ [8], have been proposed for medical image segmentation tasks, and have achieved significant advantages over the traditional methods. These advantages rely heavily on fully supervised learning that requires large-scale medical images with per-pixel/voxel annotations (Fig.1(b)). Such requirements may not be so easily met because experienced specialists are normally required to annotate each pixel/voxel laboriously. As an alternative, weakly supervised semantic segmentation is currently attracting significant attention as only low-cost partial annotations, e.g., point [9], scribble [10] or box [11] annotations, are used to train CNNs.

Training CNNs with partial annotations in the manner of supervised learning roughly by assuming that the unlabeled pixels belong to background category may confuse the models, thus leading to a foreground-suppression segmentation result as shown in Fig. 1(e); while training CNNs only with labeled pixels may generate a foreground-expansion segmentation result as depicted in Fig. 1(f). To improve segmentation quality of CNNs trained with partial annotations, many weakly supervised segmentation methods have been proposed, which can be roughly grouped into two categories, i.e., pseudo label methods [9][10][11] and regularized loss methods [12][13][14]. Typically, pseudo label methods as summarized in [15] iterate three steps: (1) pre-processing for initial pseudo labels based on region growing [10], Voronoi diagram and k-means clustering methods [9]; (2) training CNNs with pseudo labels in the manner of fully supervised learning, and (3) post-processing based on graph search [11] and DenseCRF [10] for improved pseudo labels. Such methods may be sensitive to the initial pseudo labels and perform well when pseudo labels are estimated accurately. In comparison, regularized loss methods train CNNs with partial annotations directly, but impose prior knowledge on CNNs' outputs to guide the training. The prior knowledge, e.g., the size and shape of the object of interest, is typically described as inequality constraints, and is sequentially programmed in the form of regularization terms to constrain the training process. These constraints mainly include global constraints, e.g., size constrained-CNN loss (SCCL) [14] to constrain the sizes of the object of interest, and local constraints, e.g., DenseCRF loss [12] to force CNNs' outputs to respect the pixel consistency of input image in color and spatial dimension. Regularized loss methods, i.e., constrained-CNN loss methods, are able to generate outstanding segmentation outputs when prior knowledge exists and can be described by programming language and further be embed into CNNs' training objective effectively. Unlike natural images, extensive domain-specific knowledge exists in medical image analysis, thus making regularized loss method a concise and generalized approach for weakly supervised medical image segmentation. Thus, in this paper we focus mainly on constrained-CNN loss methods.

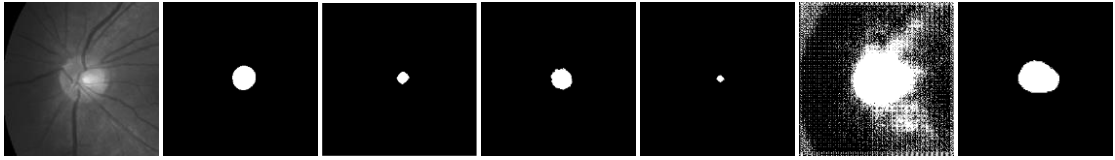

Figure 1. Illustration of segmentation results of UNet [5] trained for retinal optic cup segmentation with full supervision and partial supervision. Left-to-right: (a). Retinal fundus image. (b) Fully supervised annotations of optic cup. (c) Weakly supervised (partial) annotations of optic cup. (d) Segmentation output of a UNet trained with full supervision. (e) Segmentation output of a UNet trained with partial annotations, where the unlabeled pixels are assumed to be background pixels. (f) Segmentation output of a UNet trained with partial annotations, where only the labeled pixels are used (using partial cross-entropy loss). (g) Segmentation output of a UNet trained with size constrained-CNN loss [14].

Size constrained-CNN loss [14] is a representative work of constrained-CNN loss methods. It assumes there exists prior knowledge about the size of the object of interest, which is a common scenario in medical image segmentation [16][17]. Such size prior knowledge is described as the statistics of target size, i.e., lower bound $a$ and upper bound $b$ on size. Given training image with its weak (partial) annotations, the general way to impose size prior knowledge on CNNs' outputs can be formalized as the optimization of a partial cross-entropy loss subject to inequality constraints [18]:

$$\min_{\theta} \frac{1}{\|S\|_0} \sum_{i \in P} l(\hat{y}_i, s_i) \quad \text{s.t.} \quad a \leq \|\hat{S}\|_0 \leq b, \tag{1}$$

where $P$ denotes labeled pixels, $s_i \in \{0,1\}$ and $\hat{y}_i \in [0,1]$ is ground-truth (GT) label and CNNs' *softmax* probability output at pixel $i$, $l(\hat{y}_i, s_i) = -s_i \log(\hat{y}_i)$, $\|\hat{S}\|_0$ and $\|S\|_0$ represent L0-norm of segmentation output $\hat{S}$ and that of GT label $S$, respectively. SCCL designs a differential size constrained-CNN loss as shown in formula (2) to impose penalty on the segmentation that violates inequality constraints so as to avoid expensive Lagrangian dual iterates in training CNNs.

$$\min_{\theta} \frac{1}{\|S\|_0} \sum_{i \in P} l(\hat{y}_i, s_i) + \lambda_s \, C(\hat{S}), \tag{2}$$

where $\lambda_s$ denotes the weight of size penalty and $C(\hat{s})$ is the penalty term:

$$C(\hat{S}) = \begin{cases} (\|\hat{S}\|_0 - a)^2, & \text{if } \|\hat{S}\|_0 \leq a \\ (\|\hat{S}\|_0 - b)^2, & \text{if } \|\hat{S}\|_0 \geq b, \\ 0, & \text{otherwise} \end{cases} \tag{3}$$

Soft size, i.e., summation of *softmax* probabilities $\sum_i \hat{y}_i$, is employed instead of $\|\hat{S}\|_0$ in practice. Such penalty term is able to constrain the size of foreground area effectively to avoid excessive foreground-expansion caused by the optimization of partial cross-entropy loss as depicted in Fig. 1(f) and Fig. 1 (g). However, we observe that the segmentation output of a UNet trained with SCCL still shows a certain level of foreground-expansion. It happens because when the size of the object of interest is suppressed to the upper bound $b$, the size penalty would become zero and thus SCCL has no more energy to further constrain the size of foreground area. Such problem may get worse in the cases when the object of interest has a wide size range. This problem cannot be solved by increasing the weight of size penalty $\lambda_s$. Besides, in the paradigm of existing constrained-CNN loss methods, one need to describe the prior knowledge by a programming language, which may be difficult in a breadth of cases [15], e.g., irregular shape and unsmooth boundary of the object of interest. A new paradigm of constrained-CNN loss methods is eager to be designed to enable exploiting the rich prior knowledge in medial image analysis conveniently and effectively for weakly supervised medical image analysis.

Adversarial learning is an important technique in unsupervised domain adaptation [19], which tries to enforce CNNs to learn a domain invariant representation (e.g., deep features) for related tasks across different domains. This is done by introducing a domain discriminator $D$ to enable adversarial learning with two iterative steps: $D$ learns to distinguish which domains deep features belong to, while CNNs try to fool $D$ to maximize the probability of the source domain features being classed as target domain features. The distribution of source domain features gradually approximates that of target domain features through the iterative optimization. This process could be naturally viewed as using the prior knowledge, i.e., available images in target domain, to constrain CNNs through adversarial learning. Such constraints on distribution is obviously higher-order and strong constraints.

Inspired by adversarial learning in unsupervised domain adaptation [19], we propose adversarial constrained-CNN loss (ACCL), a new paradigm of constrained-CNN loss methods, for weakly supervised medical image segmentation. In the new paradigm, prior knowledge, e.g., the size and shape of the object of interest, is encoded and depicted by reference masks, and is further employed to impose constraints on segmentation outputs through adversarial learning with reference masks. Unlike pseudo label methods for weakly supervised segmentation, such reference masks are used to train a discriminator rather than a segmentation network, and thus are not required to be paired with specific images. Thus, the reference masks that encode the prior knowledge, e.g., the size and shape of the object of interest, can be generated easily through painting or off-the-shelf data augmentation methods, e.g., random translation, random flipping and random scaling, etc. Our new paradigm not only greatly facilitates imposing prior knowledge on network's outputs, but also provides stronger and higher-order constraints, i.e., distribution approximation, through adversarial learning. Extensive experiments involving different medical modalities, different anatomical structures, different topologies of the object of interest, different levels of prior knowledge and weakly supervised annotations with different annotation ratios is conducted to evaluate our ACCL method. Consistently superior segmentation results over SCCL have been achieved, some of which are close to the results of full supervision, thus fully verifying the effectiveness and generalization of our method. Specifically, we report an average Dice score of 75.4% with an average annotation ratio of 0.65%, surpassing the prior art, i.e., SCCL, by a large margin of 11.4%.

The significant contributions of this paper are:
(a) We propose ACCL, a new paradigm of constrained-CNN loss methods for weakly supervised medical image segmentation, which exploits adversarial learning to impose constraints of prior knowledge on segmentation outputs.
(b) In our new paradigm, imposing prior knowledge on segmentation outputs becomes much easier, more effective and more unified than the old paradigm.
(c) Our ACCL is able to provide stronger and higher-order constraints by approximating distribution.
(d) We conduct extensive experiments involving different medical modalities, different anatomical structures, different topologies of the object of interest, different levels of prior knowledge and weakly supervised annotations with different annotation ratios to evaluate the effectiveness and generalization ability of our method. Consistently superior segmentation results over SCCL have been achieved, some of which are close to the results of full supervision. Thus, our ACCL has the potential to close the gap between weakly and fully supervised learning in semantic medical image segmentation.

## 2. Adversarial constrained-CNN loss

Inspired by adversarial learning in unsupervised domain adaptation [19], we propose a new paradigm of imposing prior knowledge on CNNs' outputs for weakly supervised medical image segmentation, i.e., adversarial constrained-CNN loss, by introducing a discriminator $D$ to training procedure as diagrammed in Fig 2. In our paradigm, prior knowledge is encoded by inference masks. The discriminator $D$ is trained to distinguish whether input masks satisfy the constraints of prior knowledge underlying the distribution of inference masks. The segmentation network $G$ is tasked to not only fool the discriminator but also to fit weak annotations. Finally, $G$ learns to produce plausible segmentation gradually through the iterative optimization.

### 2.1. Reference masks

In our paradigm, prior knowledge, e.g., the size and shape of the object of interest, is encoded and depicted by reference masks. Compared with the paradigm of existing constrained-CNN loss methods, depicting prior knowledge by reference masks is much easier than describing it in programming language.

For instance, one can depict the size, irregular shape and unsmooth boundary of the object of interest easily through painting. More reference masks that satisfy such constraints can be generated by off-the-shelf data augmentation methods, e.g., random translation, random flipping, random warping and random scaling, etc. Unlike pseudo label methods, we use these reference masks to train the discriminator instead of training segmentation network. Therefore, it is not necessary to make reference masks be paired with specific training images. It should be noted that this is an important property of our paradigm, which facilitates the encoding of prior knowledge into reference masks and facilitates the usage of these reference masks in the training procedure. In our experiments, we use three different kinds of reference masks, including partial annotations, unpaired GT annotations and paired GT annotations.

**Partial ACCL**. The partial annotations, where the unlabeled pixels are assumed to be background pixels, can be used free as less accurate reference masks that encode less prior knowledge. We suppose that the partial annotations still preserve the topology of the object of interest, although its size and shape might be incorrect as shown in Fig. 1(b) and Fig .1(c). Therefore, such reference masks can be adopted by ACCL to suppress the foreground-expansion caused by the optimization of partial cross-entropy. We refer to it as partial ACCL. Partial ACCL can be viewed as the lower bound of our ACCL method because no other prior knowledge is encoded. In our experiments, we demonstrate that tuning the value of the weight of partial ACCL during the training of a segmentation network can force the network's outputs to vary from foreground-expansion to foreground-suppression. An appropriate value of the weight of partial ACCL can be determined during this period.

**Unpaired ACCL**. The unpaired GT annotations are randomly sampled from fully supervised annotations and are used as accurate reference masks that encode the statistics of object's size and shape only. We call ACCL that uses unpaired GT annotations unpaired ACCL, a representative of ACCL methods. Compared with paired GT annotations, the unpaired GT annotations does not have the pixel-level matchup with images, and thus contain only the global prior knowledge about the size and shape of the object of interest. Therefore, unpaired ACCL is expected to impose such constraints of prior knowledge on network's outputs. In practice, one can easily generate reference masks that satisfy the constraints on size and shape of the object of interest through painting and data augmentation methods described above.

**Paired ACCL**. ACCL that uses paired GT annotations, i.e., fully supervised annotations, as the reference masks is named by paired ACCL. Paired ACCL may be the upper bound of ACCL methods, because the fully supervised annotations are the most accurate and the best reference masks theoretically for ACCL. In spite of no availability in practice, we can employ it to test the upper bound on segmentation performance for our proposed ACCL method.

The reference masks are expected to guide the segmentation network $G$ to generate outputs that approximates the distribution of reference masks through adversarial training to satisfy the constraints of prior knowledge.

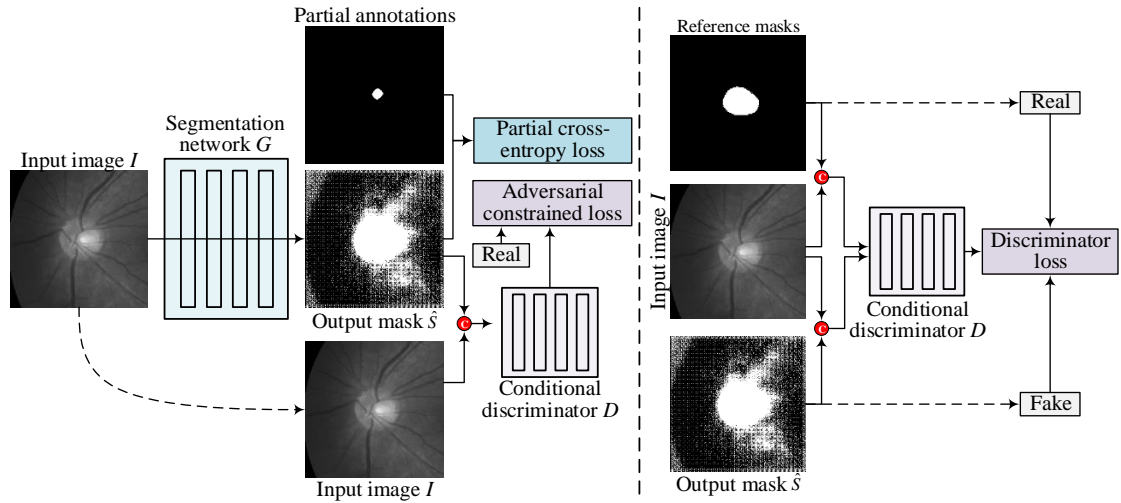

Figure 2. Training procedure of using adversarial constrained-CNN loss for weakly supervised medical image segmentation.

**2.2. Adversarial constrained-CNN loss**.

Adversarial learning is commonly used in generative models to enforce distribution approximation, which can be viewed as higher-order and stronger constraints in comparison of other constraints, e.g., size and shape constraints of target regions. Thus, we introduce a discriminator $D$ to be trained with reference masks to impose constraints of prior knowledge on segmentation outputs. Moreover, we observe that the UNet [5] trained only with partial cross-entropy loss tends to produce segmentation outputs with high-frequency noise in local patches as shown in Fig. 1(f). Based on this observation, we choose to use a *Patch*GAN-style [20] discriminator that is conditioned on input image to impose constraints at image patch scale. Besides, the objective of LSGAN [21] is adopted as our objective to train $G$ and $D$. Following the training procedure of generative adversarial network (GAN), the segmentation network $G$ and discriminator $D$ are alternatively optimized in the form of adversarial learning. Specifically, the segmentation network $G$ is tasked to not only fit partial GT annotations but also to fool the discriminator. The objective of $G$ can be expressed as:

$$\min_{\theta_g} \frac{1}{\|S\|_0} \sum_{i \in P} l(\hat{y}_i, s_i) + \lambda_a L_{ACCL}(X, \hat{Y}), \tag{4}$$

where $\lambda_a$ denotes the weights of ACCL and $L_{ACCL}(X, \hat{Y})$ is formalized as:

$$L_{ACCL}(X, \hat{Y}) = \frac{1}{M} \sum_{i=1}^{M} (D(X_i, G(X)_i) - 1)^2, \tag{5}$$

where $M$ denotes the number of $N \times N$ patches and $X_i$ is $i$-th patch of input image. The weights of ACCL $\lambda_a$ can be viewed as a confidence that describes how accuracy these reference masks encode prior knowledge. Thus, a large $\lambda_a$ is preferred when the reference masks encode prior knowledge accurately, e.g., unpaired or paired GT annotations; otherwise, a small $\lambda_a$ is preferred, e.g., partial annotations. The discriminator $D$ is trained to distinguish whether input masks satisfy the constraints of prior knowledge underlying the distribution of inference masks. Thus, we optimize this objective:

$$\min_{\theta_d} \frac{1}{M} \sum_{i=1}^{M} (D(X_i, G(X)_i))^2 + \frac{1}{M} \sum_{i=1}^{M} (D(X_i, R_i) - 1)^2, \tag{6}$$

where $R_i$ denotes $i$-th patch of reference mask $R$. Finally, $G$ learns gradually to satisfy the constraints of prior knowledge through adversarial learning and thus produce plausible segmentation.

### 3. Experiments and Results

Extensive experiments are conducted on three medical image datasets with different modalities and anatomic structures, involving retinal fundus images for optic disc/cup segmentation in the 2018 REFUGE challenge[1], our own ultrasound scans used for pupil segmentation, and magnetic resonance (MR) exams for cardiac multi-structures segmentation in the 2017 ACDC challenge[2], to validate the effectiveness of the proposed method. Specifically, we introduce three different ACCL models, i.e., partial ACCL, unpaired ACCL and paired ACCL, and four baseline models, i.e., weak CE, partial CE, SCCL and FS CE, where the differences between these models are the levels of supervision.

**Weak CE**. Weak CE trains a segmentation network with cross-entropy loss directly using weakly supervised annotations, where the unlabeled pixels are roughly assumed to be background pixels.

**Partial CE**. Partial CE trains a segmentation network with partial cross-entropy loss, where only the labeled pixels are used.

**SCCL**. SCCL trains a segmentation network with partial cross-entropy loss and size constrained-CNN loss (defined by Eqs. (3)). SCCL is a representative work of constrained-CNN loss methods. We use SCCL to impose size constraint on segmentation outputs.

**FS CE**. FS CE trains a segmentation network with cross-entropy loss using fully supervised annotations and serve as a measure of the performance of weakly supervised segmentation.

**Partial ACCL**. Partial ACCL uses weakly supervised annotations as the reference masks of ACCL to train a segmentation network with partial cross-entropy loss and ACCL. We regard partial ACCL as a lower bound of ACCL methods and explore how ACCL works in the lower bound.

**Unpaired ACCL**. Unpaired ACCL uses unpaired GT annotations as the reference masks of ACCL to train a segmentation network with partial cross-entropy loss and ACCL. The unpaired GT annotations satisfy the size, shape and boundary constraints of prior knowledge. We employ it to simulate the reference masks generated by painting and data augmentation under the guidance of prior knowledge.

**Paired ACCL**. Paired ACCL uses paired GT annotations as the reference masks of ACCL to train a segmentation network with partial cross-entropy loss and ACCL. We regard paired ACCL as an upper bound of ACCL methods and explore how ACCL works in the upper bound.

We train these models on the training sets of the datasets described above and test them on the test sets. Besides, we specially explore and discuss the effect of $\lambda_a$ in Eqs. (4), i.e., the weight of ACCL, by adjusting the value of $\lambda_a$ when training ACCL models on the ACDC dataset.

### 3.1 Overall training and implementation details

For all the experiments, we employ UNet [5] with 8 down sampling layers, single channel input layer with and single channel output layer as the segmentation network $G$. Since UNet is one of the most successful segmentation framework in medical imaging, we expect that the results can easily generalize to other medical image analysis tasks. The conditional discriminator of *Patch*GAN in [20] is adopted to implement the proposed ACCL. The weight of SCCL $\lambda_s$ is set to 0.01 according to [14]. UNet and the discriminator $D$ are trained from scratch by using Adam optimizer with the parameters of $\beta_1 = 0.5$ and $\beta_2 = 0.999$. We train the models with 200 epochs using mini-batches of 1. We use an initial learning rate of 2e-4 that is linearly decayed by 1% each epoch after 100 epochs. The common Dice similarity coefficient (DSC) is computed to compare the different models. We implement our code with PyTorch and NumPy, and run all the experiments on a laptop machine equipped with a NVIDIA GTX1060 (notebook) GPU (6 GBs of memory).

---
[1] https://refuge.grand-challenge.org/
[2] https://www.creatis.insa-lyon.fr/Challenge/acdc/

Table 1. Segmentation results of different models. We use the mean DSC to measure the segmentation quality, where a larger value means a better segmentation results. The annotation ratio denotes the percentage of labeled pixels in the weakly supervised annotations. We compute the average DSC score on all the four tasks as the overall performance.

| Anatomical structure | Modality | Topology | Annotation ratio | Weak CE | Partial CE | SCCL | Partial ACCL | Unpaired ACCL | Paired ACCL | FS CE |
|---|---|---|---|---|---|---|---|---|---|---|
| Optic Disc | Retinal fundus image | *ringlike* | 1.92% | 45.2 | 23.0 | 76.7 | 78.5 | 82.3 | 83.2 | 86.6 |
| Optic Cup | Retinal fundus image | *globular* | 0.52% | 35.1 | 10.6 | 68.0 | 73.0 | 81.1 | 80.9 | 84.0 |
| Ventricular Endocardium | MR slice | *globular* | 0.12% | 7.5 | 4.8 | 57.5 | 59.9 | 63.4 | 66.2 | 81.2 |
| Pupil | Ultrasound image | *globular* | 0.05% | 9.2 | 1.3 | 53.8 | 72.7 | 74.8 | 75.9 | 92.7 |
| Overall | | | 0.65% | 24.3 | 9.9 | 64.0 | 71.0 | 75.4 | 76.55 | 86.1 |

**3.2 Experiments on weakly supervised optic disc and cup segmentation**

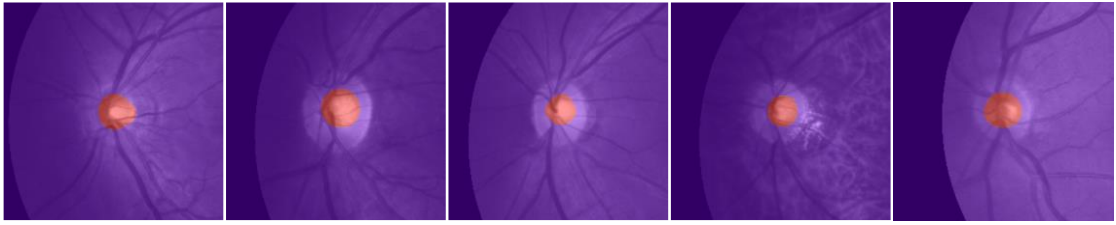

(a) Fully supervised annotations

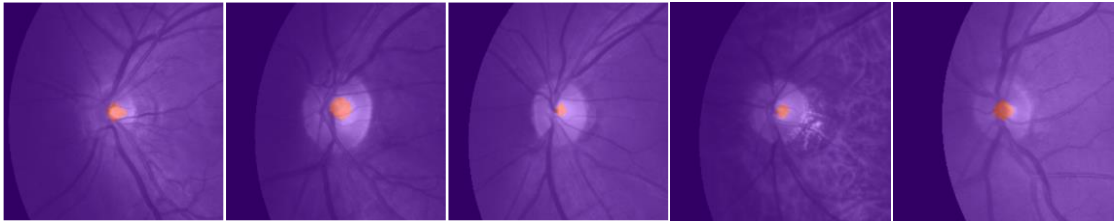

(b) Weakly supervised annotations

Figure 3. Examples of fully supervised annotations and weakly supervised annotations in REFUGEcup dataset. In the fully labeled images (top), all pixels are annotated, with purple depicting the background and red depicting the object of interest. In the weakly supervised cases (bottom), only the labels of the red pixels are known. The average percentage of labeled pixels in weakly supervised annotations is 0.52%.

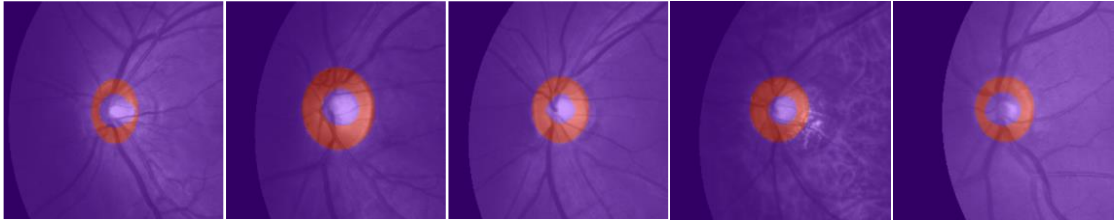

(a) Fully supervised annotations

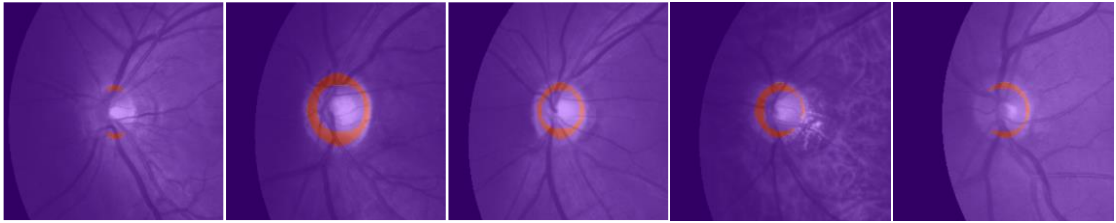

(b) Weakly supervised annotations

Figure 4. Examples of fully supervised annotations and weakly supervised annotations in REFUGEdisc dataset. In the fully labeled images (top), all pixels are annotated, with purple depicting the background and red depicting the object of interest. In the weakly supervised cases (bottom), only the labels of the red pixels are known. The average percentage of labeled pixels in weakly supervised annotations is 1.92%.

**REFUGEcup and REFUGEdisc dataset.** The REFUGE dataset consists of 400 retinal fundus images with size $2124 \times 2056$ acquired by a Zeiss Visucam 500 camera for training. We first center and crop optic disc regions with a size of $1000 \times 1000$ and then resize these images to a small size of $256 \times 256$ in order to adapt the network's receptive. We split the fully supervised annotations of optic disc and optic cup to build two individual datasets, i.e., REFUGEcup and REFUGEdisc, for *ringlike* optic disc segmentation and *globular* optic cup respectively, which is able to evaluate our method on the object of interest with different topologies. Next, we randomly select 100 images, involving 75 images for training and 25 images for test, from 400 retinal fundus images to create small sample datasets. Since collecting and labeling medical images may not be so easy in most cases, evaluating our method on small sample datasets is much more useful in practice. Finally, we follow the method in [14] to generate weak (partial) labels through binary erosion on the fully annotations. As a result, the total numbers of annotated pixels of weak labels in REFUGEcup and REFUGEdisc datasets represent the 0.52% and 1.92% of the labeled pixels in the fully supervised scenario respectively as depicted in Fig. 3 and Fig. 4.

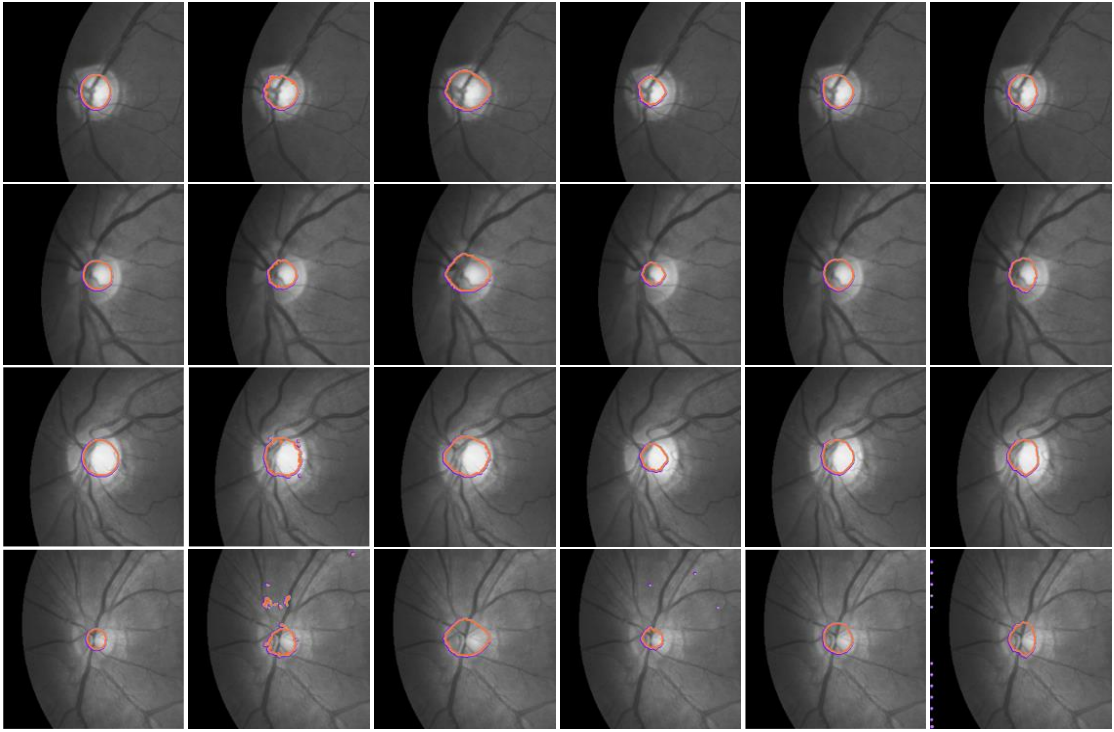

Figure 5. Qualitative comparison of the different methods on REFUGEcup dataset. Each row represents a retinal fundus image from test set and each column stands for one method. Left-to-right: (a) GT annotations; (b) FS CE; (c) SCCL; (d) partial ACCL; (e) unpaired ACCL and (f) paired ACCL. (Best viewed in colors)

**Training**. We prepare three kinds of reference masks, i.e., partial annotations, unpaired GT annotations and paired GT annotations, as described in **section 2.1** for ACCL. Meanwhile, we calculate the minimum and maximum values of the target size over 100 full GT masks with size of 256×256 in each dataset, and multiply by a factor of 0.9 and 1.1 to be used as the lower and upper bounds on target size in Eqs. (3) for SCCL. As a result, the lower and upper bounds on target size ($a$, $b$) of REFUGEcup and REFUGEdisc dataset are (508, 3947) and (1359, 5625), respectively. We empirically set the weight values of partial ACCL, unpaired ACCL and paired ACCL to 1.5e-3, 5.0e-2 and 5.0e-3 for REFUGEcup dataset, and 2.5e-3, 2.0e-2, 2.0e-2 for REFUGEdisc dataset. Other experiment settings are detailed in **Section 3.1**.

**Results and Discussion on REFUGEcup dataset**. The quantitative result of each model is shown in the third row of Table 1. First, the weak CE model and partial CE model trained without constrained-CNN loss show a poor segmentation performance; while benefiting from the constraints on segmentation outputs, these models trained with constrained-CNN loss achieve much better segmentation results. Next, we find that all these three ACCL models achieve significant higher DSC scores than SCCL. Specifically, even the partial ACCL, using only less accurate reference masks to be noted, surpasses SCCL by a large margin of 5% in mean DSC score. Moreover, with the help of accurate prior knowledge, the unpaired ACCL and paired ACCL that use ~0.5% of fully supervised annotations achieve results comparable to fully supervised FS CE model. It indicates that our ACCL may have the potential to close the gap between weakly and fully supervised learning in semantic medical image segmentation. Besides, we also notice that the unpaired ACCL surpasses partial ACCL by 8.1% in mean DSC score, and is comparable to paired ACCL. The former implies that using accurate global prior knowledge, e.g., the size range of the object of interest, can effectively improve the segmentation quality of weakly supervised learning through proposed ACCL, while the latter indicates that ACCL is not sensitive to local prior knowledge, e.g., the specific size and position of the object of interest in specific image. This property of ACCL greatly facilitate the generation of reference masks. On the other hand, we also make a qualitative comparison to explore the segmentation quality of each model as depicted in Fig. 4. We can observe that all these models trained with constrained-CNN loss are able to produce plausible segmentation. Nevertheless, the segmentation outputs of SCCL model present foreground-expansion to a certain extent (third column in

Fig. 4). Such phenomenon has verified our analysis about the SCCL's problem: when the constraints of the upper bound *b* on size is satisfied, SCCL could not provide any energy to further suppress the foreground-expansion caused by optimizing the partial cross-entropy loss. This problem cannot be solved by tuning the weight of SCCL. In comparison, our ACCL method does not have this problem, and can generate more plausible segmentation, where the target size is neither too small nor too large.

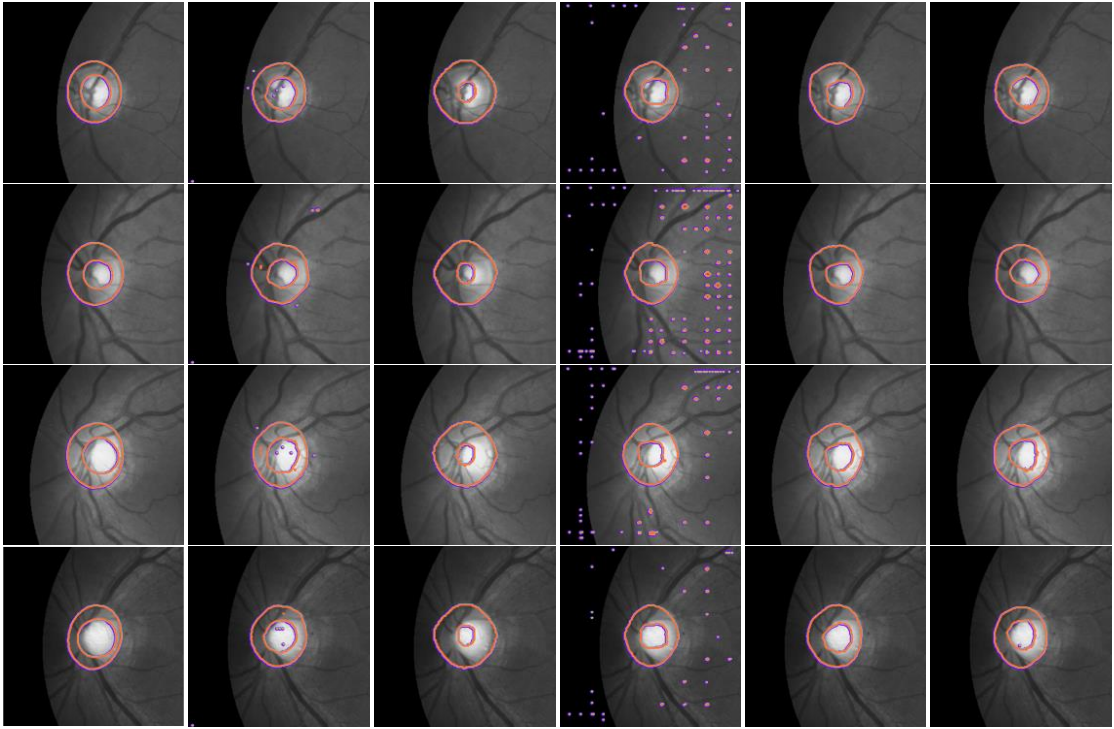

Figure 6. Qualitative comparison of the different methods on REFUGEdisc dataset. Each row represents a retinal fundus image from test set and each column stands for one method. Left-to-right: (a) GT annotations; (b) FS CE; (c) SCCL; (d) partial ACCL; (e) unpaired ACCL and (f) paired ACCL. (Best viewed in colors)

**Results and Discussion on REFUGEdisc dataset**. Unlike the *globular* optic cup, optic disc presents *ringlike*. The topologies of optic cup and optic disc are different. Thus, we can explore how constrained-CNN loss method works on the object of interest with a ring topology in this case. The quantitative results of different models are listed in the second row of Table 1, while qualitative comparison is visualized in Fig. 6. We find that our methods can learn a segmentation network with ~1.9% of fully supervised annotations to segment a *ringlike* object, and achieve consistently superior results over SCCL. Seeing from the visual results of SCCL (third column in Fig. 6), not surprisingly, we can also observe foreground-expansion problem of SCCL (the inner contour of *ringlike* object is smaller than GT). Although partial ACCL gets a higher DSC score than SCCL, it produces some small false positive regions scattered throughout the whole image (forth column in Fig. 6). We analyze that it probably because the segmentation network is misled by inaccurate prior knowledge encoded in reference masks of partial ACCL. Specifically, we notice the weakly supervised annotations shown in the first column of Fig. 4(b) does not preserve the topology of *ringlike* optic disc. The inaccurate topology might be magnified during adversarial training, thus leading to such implausible segmentation. In comparison, both unpaired ACCL and paired ACCL generate plausible segmentation and obtain good performance comparable to FS CE. Therefore, we suggest to preserve the topology of the object of interest when labelling the partial annotations.

### 3.3 Experiments on weakly supervised ultrasound pupil segmentation

**PUPIL dataset**. The PUPIL dataset consists of 100 ultrasound scans of the eyes with size of $330 \times 450$. We randomly sample 75 images for training and 25 images for test, and resize these images to the size of $256 \times 256$. We generate weakly supervised annotations in the same way as REFUGEcup dataset does. As shown in Fig. 7, the average percentage of labeled pixels in weakly supervised annotations is 0.05%, which means only about 32 pixels are labelled in each image with size of $256 \times 256$.
**Training**. We prepare three kinds of reference masks, i.e., partial annotations, unpaired GT annotations and paired GT annotations, as described in **section 2.1** for ACCL. Meanwhile, we calculate that the lower and upper bounds on target size $(a, b)$ is (306, 2053) for SCCL in the same way as described in **section 3.2**. We empirically set the weight values of partial ACCL, unpaired ACCL and paired ACCL to 1.0e-3, 5.0e-3 and 5.0e-2. Other experiment settings are detailed in **Section 3.1**.
**Results and Discussion**. The quantitative results of different models are shown in the fifth row of Table 1, and the qualitative comparison is depicted in Fig. 8. Compared with REFUGEcup and REFUGEdisc dataset, the weakly supervised annotations in PUPIL dataset has a notable lower annotation ratio (0.05%), which brings a severe challenge to weakly supervised segmentation algorithms. In this case, all these models except for FS CE (full supervision) get lower DSC scores as expected than the results on REFUGEcup and REFUGEdisc dataset. Specifically, we find that our methods achieve consistently superior results over SCCL, where even the lower bound of our methods, i.e., the partial ACCL, gets an almost 20% higher DSC score than that of SCCL. It implies that our methods can better adapt to weakly supervised scenario with low annotation ratio than SCCL. The qualitative results also demonstrate the superiority of our ACCL methods in producing plausible segmentations.

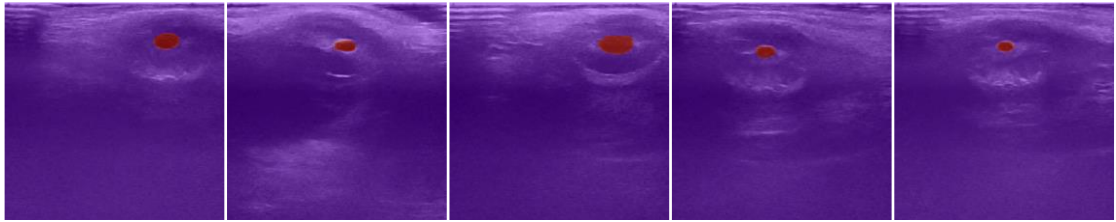

(a) Fully supervised annotations

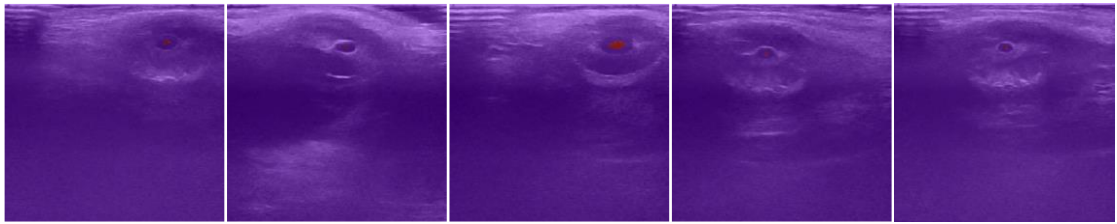

(b) Weakly supervised annotations

Figure 7. Examples of fully supervised annotations and weakly supervised annotations in PUPIL dataset. In the fully labeled images (top), all pixels are annotated, with purple depicting the background and red depicting the object of interest. In the weakly supervised cases (bottom), only the labels of the red pixels are known. The average percentage of labeled pixels in weakly supervised annotations is 0.05%.

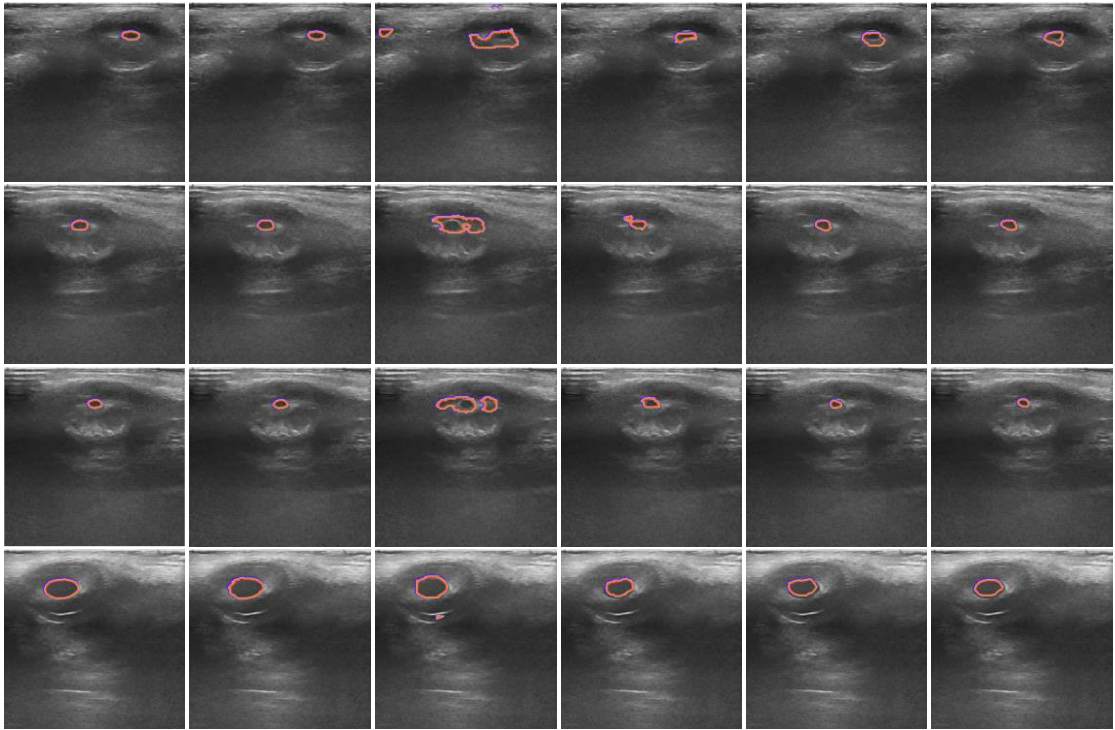

Figure 8. Qualitative comparison of the different methods on PUPIL dataset. Each row represents a ultrasound scan from test set and each column stands for one method. Left-to-right: (a) GT annotations; (b) FS CE; (c) SCCL; (d) partial ACCL; (e) unpaired ACCL and (f) paired ACCL. (Best viewed in colors)

### 3.4 Experiments on weakly supervised left ventricular endocardium segmentation

**ACDC dataset**. This dataset includes 100 cine MR exams covering well defined pathologies: dilated cardiomyopathy, hypertrophic cardiomyopathy, myocardial infarction with altered left ventricular ejection fraction and abnormal right ventricle [22]. In our experiments, the object of interest is set as the left ventricular endocardium. We randomly select 75 exams, slice them into slices and resize these slices to size of $256\times 256$. Subsequently, we randomly sample 75 slices that contain the object of interest for training. We do same operations on the remaining 25 exams and thus get 25 slices for test. At last, we generate weakly supervised annotations in the same way as REFUGEcup dataset does. As shown in Fig. 9, the average percentage of labeled pixels in weakly supervised annotations is 0.12%, which means only about 82 pixels are labelled in each image with size of $256\times 256$.

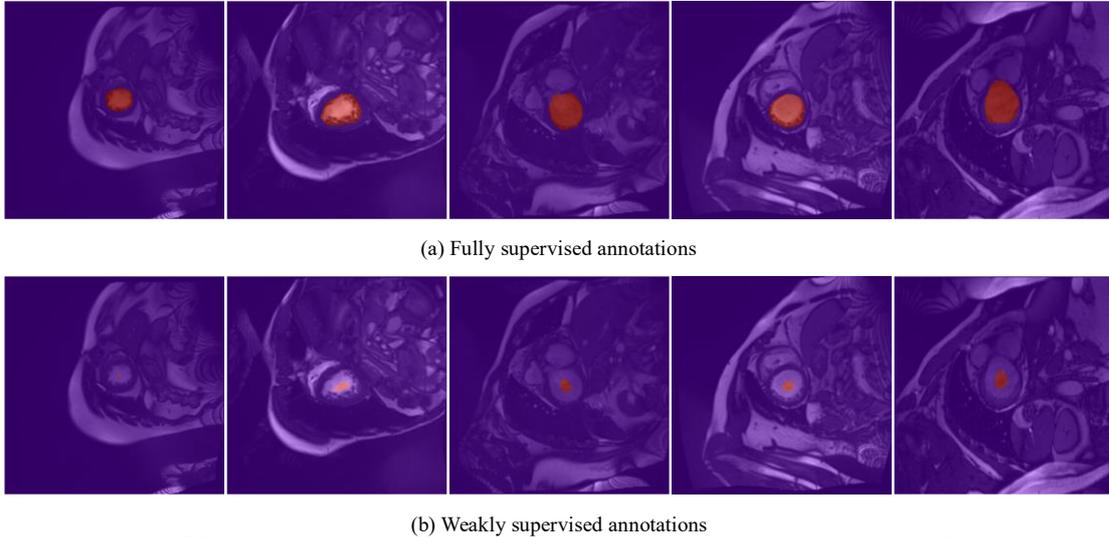

(a) Fully supervised annotations

(b) Weakly supervised annotations

Figure 9. Examples of fully supervised annotations and weakly supervised annotations in ACDC dataset. In the fully labeled images (top), all pixels are annotated, with purple depicting the background and red depicting the object of interest. In the weakly supervised cases (bottom), only the labels of the red pixels are known. The average percentage of labeled pixels in weakly supervised annotations is 0.12%.

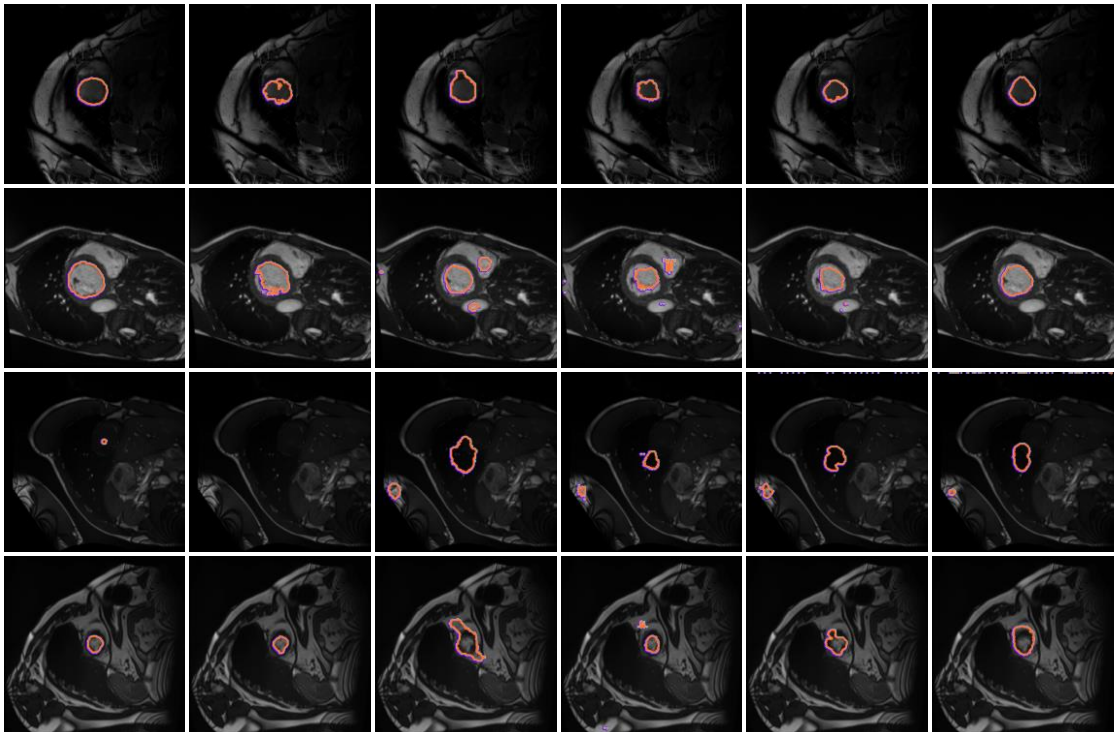

Figure 10. Qualitative comparison of the different methods on ACDC dataset. Each row represents a cardiac MR slice from test set and each column stands for one method. Left-to-right: (a) GT annotations; (b) FS CE; (c) SCCL; (d) partial ACCL; (e) unpaired ACCL and (f) paired ACCL. (Best viewed in colors)

**Training**. We prepare three kinds of reference masks, i.e., partial annotations, unpaired GT annotations and paired GT annotations, as described in **section 2.1** for ACCL. Meanwhile, we calculate that the lower and upper bounds on target size ($a$, $b$) is (23, 2333) for SCCL in the same way as described in **section 3.2**. We empirically set the weight values of partial ACCL, unpaired ACCL and paired ACCL to 6.0e-4,

5.0e-2 and 5.0e-2. Other experiment settings are detailed in **Section 3.1**. Besides, we specially explore and discuss the effect of $\lambda_a$ in Eqs. (4), i.e., the weight of ACCL, by tuning the value of $\lambda_a$. Specifically, we construct a candidate value set: {3.0e-4, 6.0e-4, 1.0e-3, 2.0e-3} for partial ACCL, {5.0e-3, 1.0e-2, 2.0e-2, 5.0e-2} for unpaired ACCL, and {5.0e-3, 1.0e-2, 2.0e-2, 5.0e-2} for paired ACCL.

**Results and Discussion**. The quantitative results of different models are shown in the forth row of Table 1, and the qualitative comparison is depicted in Fig. 10. Compared with REFUGEcup/disc and PUPIL datasets, the images in ACDC dataset have more complicated background and varied foreground, which poses a serious challenge to segmentation algorithms. In this case, our methods are still able to achieve consistently superior results over SCCL, which implies that our methods can easily adapt to complicated medical image datasets. Once again, the qualitative results demonstrate the superiority of our ACCL methods in producing plausible segmentations. On the other hand, we train partial ACCL, unpaired ACCL and paired ACCL with different $\lambda_a$ in their individual candidate value set, respectively. Partial ACCL get DSC scores of {0.542, 0.599, 0.548, 0.434} and the corresponding segmentation outputs are visualized in Fig. 11; unpaired ACCL achieves DSC scores of {0.609, 0.612, 0.613, 0.634} and the corresponding segmentation outputs are depicted in Fig. 12; paired ACCL obtains DSC scores of {0.630, 0.593, 0.618, 0.662} and the corresponding segmentation outputs are shown in Fig. 13. We notice that tuning $\lambda_a$ of partial ACCL in a small range will notable affect the performance of segmentation. Such impact can be seen more intuitively in the specific segmentation output (Fig. 11). We find that a larger $\lambda_a$ of partial ACCL is prone to obtain a foreground-suppression segmentation, while a smaller $\lambda_a$ is prone to get a foreground-expansion segmentation. We analyze that a smaller $\lambda_a$ would make the optimization of partial cross-entropy loss dominate the training and a larger $\lambda_a$ would make the optimization of ACCL dominate the training. The former tends to produce foreground-expansion segmentation, while the latter could suppress the outputs to satisfy the distribution of weakly supervised annotations. Therefore, partial ACCL is sensitive to the value of $\lambda_a$, and we can only empirically set this hyper-parameter $\lambda_a$; in general, 1.0e-3 is preferred as a good start for tuning $\lambda_a$. In comparison, unpaired and paired ACCL are less sensitive to the weight of ACCL: with a wide range of weight values, they achieve consistently superior results over SCCL; meanwhile, there are no significant differences between the segmentation results with different values of $\lambda_a$. We analyze that increasing the value of $\lambda_a$ would not lead to notable foreground-suppression as happened in partial ACCL. Therefore, we suggest that the weight of ACCL $\lambda_a$ should be set according to the reference masks of ACCL: a larger value of $\lambda_a$ is preferred, e.g., 1e-2, when reference masks, e.g., unpaired GT annotations, are able to encode accurate size and shape prior knowledge.

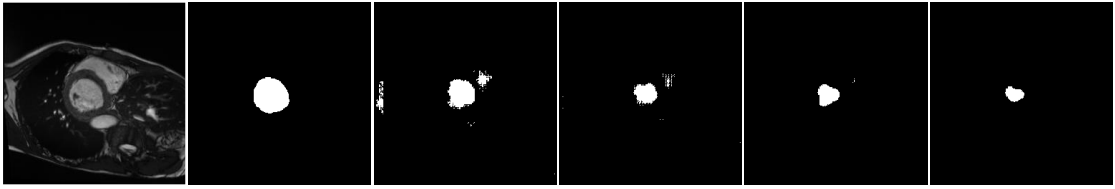

Figure 11. Illustration of segmentation results of partial ACCL with different weight values of ACCL. Left-to-right: (a) cardiac MR slice; (b) fully supervised annotations; (c), (d), (e) and (f) represent the segmentations of partial ACCL with weight values of {3.0e-4, 6.0e-4, 1.0e-3, 2.0e-3}, respectively. We can easily observe that a high weight value of ACCL is prone to obtain a foreground-suppression segmentation. Thus, we argue that partial ACCL is sensitive to the weight of ACCL.

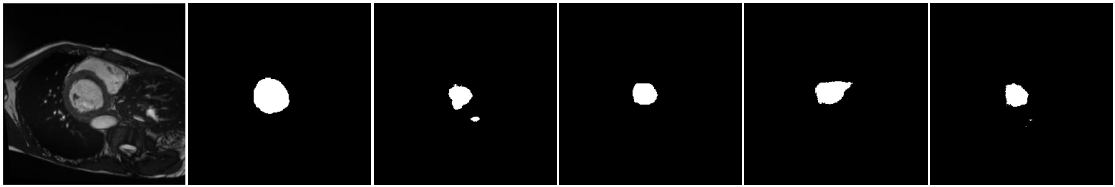

Figure 12. Illustration of segmentation results of unpaired ACCL with different weight values of ACCL. Left-to-right: (a) cardiac MR slice; (b) fully supervised annotations; (c), (d), (e) and (f) represent the segmentations of unpaired ACCL with weight values of {5.0e-3, 1.0e-2, 2.0e-2, 5.0e-2}, respectively. We cannot easily distinguish the differences between these segmentation results. Thus, we argue that unpaired ACCL is less sensitive to the weight of ACCL.

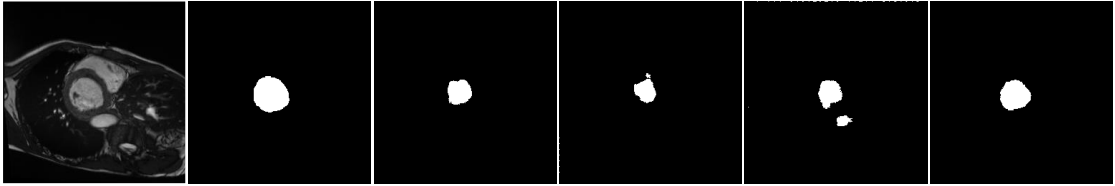

Figure 13. Illustration of segmentation results of paired ACCL with different weight values of ACCL. Left-to-right: (a) cardiac MR slice; (b) fully supervised annotations; (c), (d), (e) and (f) represent the segmentations of paired ACCL with weight values of {5.0e-3, 1.0e-2, 2.0e-2, 5.0e-2}, respectively. We cannot easily distinguish good segmentations from bad ones. Thus, we argue that paired ACCL is less sensitive to the weight of ACCL.

**Conclusion**

In this paper, we propose adversarial constrained-CNN loss, a new paradigm of constrained-CNN loss methods, for weakly supervised medical image segmentation. In the new paradigm, prior knowledge, e.g., the size and shape of the object of interest, is encoded and depicted by reference masks, and is further employed to impose constraints on segmentation outputs through adversarial learning with reference masks. Unlike pseudo label methods for weakly supervised segmentation, such reference masks are used to train a discriminator rather than a segmentation network, and thus are not required to be paired with specific images. Our new paradigm not only greatly facilitates imposing prior knowledge on network's outputs, but also provides stronger and higher-order constraints, i.e., distribution approximation, through adversarial learning. Extensive experiments involving different medical modalities, different anatomical structures, different topologies of the object of interest, different levels of prior knowledge and weakly supervised annotations with different annotation ratios have been conducted to evaluate our ACCL method. Consistently superior segmentation results over the size constrained-CNN loss method have been achieved, some of which are close to the results of full supervision, thus fully verifying the effectiveness and generalization of our method. Thus, our ACCL has the potential to close the gap between weakly and fully supervised learning in semantic medical image segmentation.

**Inference**